\documentclass[letterpaper]{article}
\usepackage{uai2019}
\usepackage[margin=1in]{geometry}

\usepackage{times}
\usepackage[square,sort,comma,numbers]{natbib}

\usepackage[utf8]{inputenc} 
\usepackage[T1]{fontenc}    
\usepackage{hyperref}       
\usepackage{url}            
\usepackage{booktabs}       
\usepackage{amsfonts}       
\usepackage{amsmath}
\usepackage{bbold}
\usepackage[dvipsnames]{xcolor}
\usepackage{nicefrac}       
\usepackage{microtype}      
\usepackage{subcaption}
\usepackage{graphicx}
\usepackage{color}

\title{Countdown Regression: Sharp and Calibrated Survival Predictions}

\author{  Anand Avati, \quad Tony Duan, \quad Sharon Zhou, \quad Kenneth Jung,  \quad Nigam H. Shah, \quad Andrew Y. Ng \\ Stanford University\\  \texttt{\{avati,tonyduan,sharonz,ang\}@cs.stanford.edu,\{kjung,nigam\}@stanford.edu} \\}
\begin{document}

\maketitle

\begin{abstract}
Probabilistic survival predictions from models trained with Maximum Likelihood Estimation (MLE) can have high, and sometimes unacceptably high variance.  The field of meteorology, where the paradigm of maximizing sharpness subject to calibration is popular, has addressed this problem  by using scoring rules beyond MLE, such as the Continuous Ranked Probability Score (CRPS). In this paper we present the \emph{Survival-CRPS}, a generalization of the CRPS to the survival prediction setting, with right-censored and interval-censored variants. We evaluate our ideas on the mortality prediction task using two different Electronic Health Record (EHR) data sets (STARR and MIMIC-III) covering millions of patients, with suitable deep neural network architectures: a Recurrent Neural Network (RNN) for STARR and a Fully Connected Network (FCN) for MIMIC-III. We compare results between the two scoring rules while keeping the network architecture and data fixed, and show that models trained with Survival-CRPS result in sharper predictive distributions compared to those trained by MLE, while still maintaining calibration.
\end{abstract}

\section{\textsc{Introduction}}

Accurate and confident predictions of the time to an event, such as patient mortality or customer churn, allow for better decision making. Methods have been developed in the survival analysis literature to address this problem of predicting time to events given censored data; that is, when sometimes we only know that an event did not happen until a certain period, or when we know an event happened in a (wide) time window, but not the exact moment. The most common approach for fitting survival models is via Maximum Likelihood Estimation (MLE) \cite{harrell_rms}, or maximum partial likelihood estimation such as in Cox Regression \cite{Cox1972}. However, MLE is equivalent to the logarithmic scoring rule, which is known to be subject to hypersensitivity \cite{Selten98,gebetsberger_estimation_2018}. The intuition is that forecasts are encouraged to be under-confident since the observation of an event that has low predicted density results in an extremely high loss. This can result in models that are overly conservative, and predict distributions with very high variance -- sometimes too high to be practically useful. Hypersensitivity of MLE is generally not a problem in typical homoskedastic regression tasks where the models only output point estimates, and uniform variance is assumed across examples. It is also not a problem in applications where the goal is accurate ranking (for example, in risk stratification). However, in the heteroskedastic regression task, where the model outputs a full probability distribution (such as a patient specific survival curve) over the outcome, the shortcomings of MLE can become a problem for practical use.



\begin{figure*}[t]
    \centering
    \includegraphics[width=\linewidth]{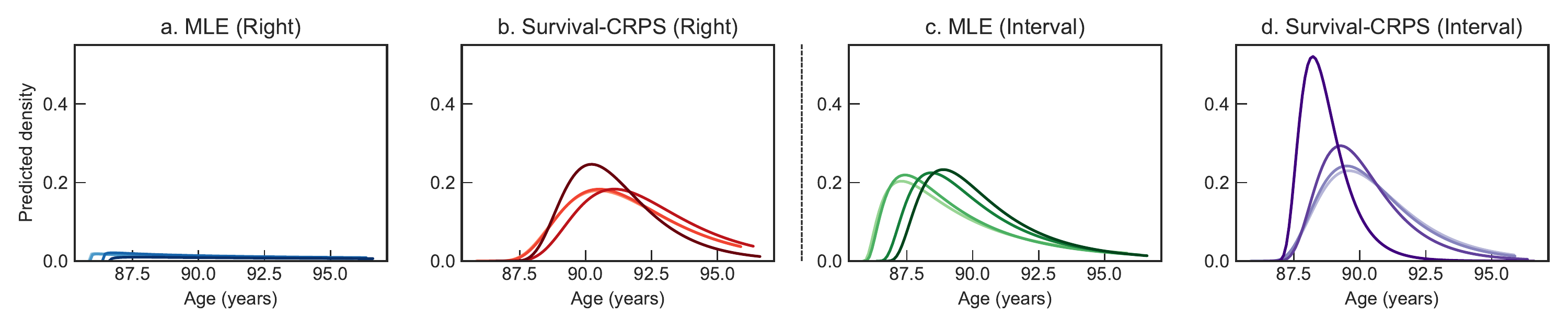}
    \caption{Example of a patient's predicted distributions for age of mortality under different objectives (losses). Repeated interactions are indicated by darker color. Our proposed techniques (b) and (d) improve sharpness of predicted distributions, while maintaining calibration from the original MLE methods (a) and (c). Graphs (a) and (b) compare MLE vs. CRPS in the right-censored setting and (c) vs. (d) in interval-censored setting (sharper the curves the better). The densities of (a) show that predicted variance from a model trained with MLE can be unacceptably high, especially in the heavily censored setting.}
    \label{fig:main_fig}
\end{figure*}

Having access to accurate and confident probabilistic predictions can especially be helpful in healthcare. Historically, a variety of prognosis scores have been developed as tools to stratify patient risk. Such scores output a single numeric number, which is ideally suited for ranking, triaging, and prioritizing care \citep{db9910e0f7a64a75866d354a536fece3, PMID:25613983, yu_learning_2011}. Naturally, metrics such as C-statistic  \citep{uno_evaluating_2007} are appropriate to evaluate ranking tools, and log-$\ell_1$ loss \citep{yu_learning_2011}, and mean-squared-error \citep{katzman_deepsurv:_2018} for point predictions. However, accurate and confident prognosis of a particular patient's future outcomes (as opposed to relative risk of this patient against others in a group) requires more nuanced forecasting of calibrated patient specific survival curves \citep{goff_2013_2014, ranganath_deep_2016, lee_deephit:_2018}. For example, when making a clinical decision of whether a given patient is at risk of a heart attack in the next 3 months vs the next 12 months, global metrics such as concordance are irrelevant, and sharp and calibrated predictions are needed instead \citep{avati_improving_2017, rajkomar_scalable_2018, sherman_leveraging_2017}.

%

\newpage
Calibration means that predicted probabilities over events match real-world frequencies of their occurrence. For example, among all the days that had a rain forecast of 80\%, it should actually rain approximately 8 of 10 time. Probabilistic forecasts are accurate if they are well calibrated. However, calibration alone is not sufficient. Hypothetically, a model could always output the marginal distribution over outcomes as its prediction, making it well calibrated, but with little practical use. This is the notion captured by the paradigm of maximizing sharpness (i.e, concentration of probability) subject to calibration, which is widely and successfully adopted in meteorology \citep{gneiting_probabilistic_2014}. The intuition behind this paradigm is that (i) uncalibrated predictions regardless of sharpness are wrong and useless,  (ii) calibrated but non-sharp predictions are correct but less useful, and (iii) calibrated and sharp distributions are most useful (the sharper the better) \citep{gneiting_probabilistic_2014}. For example, suppose a doctor is presented with predictions from multiple models, such as those shown in Figure \ref{fig:main_fig}. Although they all come from well calibrated models, it is clear that predictions from the model behind sharper distributions is more useful. Lack of sharpness is commonly encountered when training survival prediction models with MLE in the datasets that observe heavy censoring. Thus, naively using MLE based survival prediction models on large scale real-world data (where censored data is generally the majority) can be challenging. 



The field of meteorology has improved the sharpness of probabilistic forecasts by abandoning the logarithmic scoring rule, and instead using the Continuous Ranked Probability Score (CRPS). The CRPS is a robust scoring rule which is not swayed by outliers as heavily as MLE. However CRPS is not as simple (neither analytically nor numerically) as MLE. Though CRPS has been used in the regression context to train models \citep{gneiting_probabilistic_2014, mohammadi_meta-heuristic_2016, mohammadi_optimization_2015}, it does not handle censored observations, which is crucial to building survival prediction models.

\emph{Summary of contributions}: 

(i) We introduce Survival-CRPS, a generalization of CRPS to handle right and interval-censored data (Section \ref{survcrps}).

(ii) We propose a new evaluation metric, Survival-AUPRC, a generalization of the Area Under the Precision-Recall Curve that holistically measures sharpness and calibration, handling right-censored and interval-censored outcomes (Section \ref{survauprc}).

(iii) We demonstrate the benefits of Survival-CRPS  over MLE by producing sharper, and calibrated survival predictions of patient mortality on two large-scale EHR data sets (Section \ref{experiments}).

(iv) We provide practical recommendations and choices for implementing models with Survival-CRPS on large scale data (Section \ref{recommendations}).

Though we frequently use the healthcare setting  to describe ideas, all the concepts we present in this paper are completely general and apply to any right or interval censored survival prediction problem.

\section{\textsc{Countdown Regression}}

We consider a dataset of time-to-event records $\left\{ x^{(i)}, y^{(i)}, c^{(i)}, \mathcal{T}^{(i)} \right\}$, where $x^{(i)} \in \mathbb{R}^d$ denotes a set of features, $y^{(i)} \in \mathbb{R}_+$ denotes time to event or censorship, $c^{(i)} \in \{0,1\}$ is a censoring indicator where $c^{(i)} = 0$ means time to event is $y^{(i)}$, and $c^{(i)} = 1$ means time to event is at least $y^{(i)}$, and $\mathcal{T}^{(i)}$ denotes time by which the event must have occurred, where $\mathcal{T}^{(i)} = \infty$ in the right-censored setting and $\mathcal{T}^{(i)} \in \mathbb{R}_+$ in the interval-censored setting. We omit superscripts $i$ for succinctness where possible in this section.


\newpage
Parametric survival prediction methods model the time to an event of interest with a family of probability distributions, indexed by the distribution parameters. The survival function, denoted $S(t): [0, \infty) \to [0, 1]$, is a monotonically decreasing function over the positive reals with $S(0) = 1$ and $\lim_{t\to\infty} S(t) = 0$. The survival function represents the probability of an event of interest not occurring up to a given time. Every survival function has a corresponding cumulative density function (CDF), denoted $F(t) = 1 - S(t)$, and probability density function (PDF), denoted $f(t) = \frac{d}{dt}F(t)$. The choice of the family of probability distributions implies assumptions made about the nature of the data generating process.

\subsection{\textsc{Survival-CRPS: Proper Scoring Rule Objective}}\label{survcrps}

\begin{figure*}[t]
\centering
\begin{subfigure}[h]{0.3 \linewidth}
\caption{Uncensored}
\label{fig:crps-a}
\includegraphics[width=\linewidth]{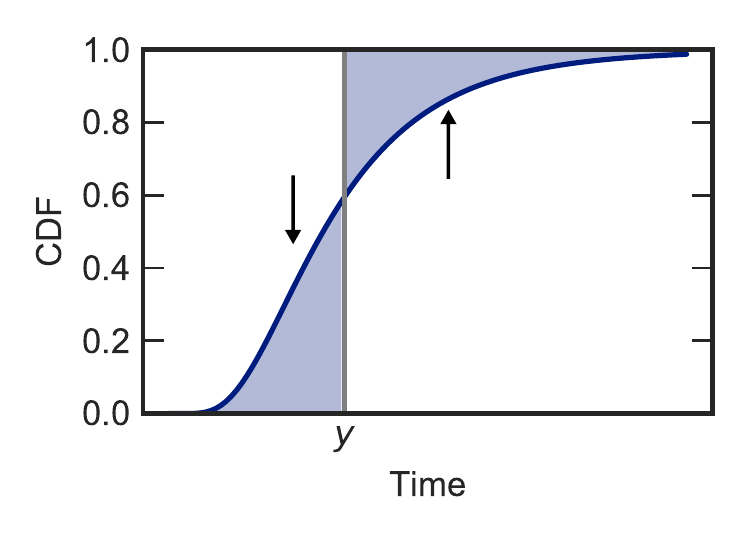}
\end{subfigure}
\begin{subfigure}[h]{0.3 \linewidth}
\caption{Right-censored}
\label{fig:crps-b}
\includegraphics[width=\linewidth]{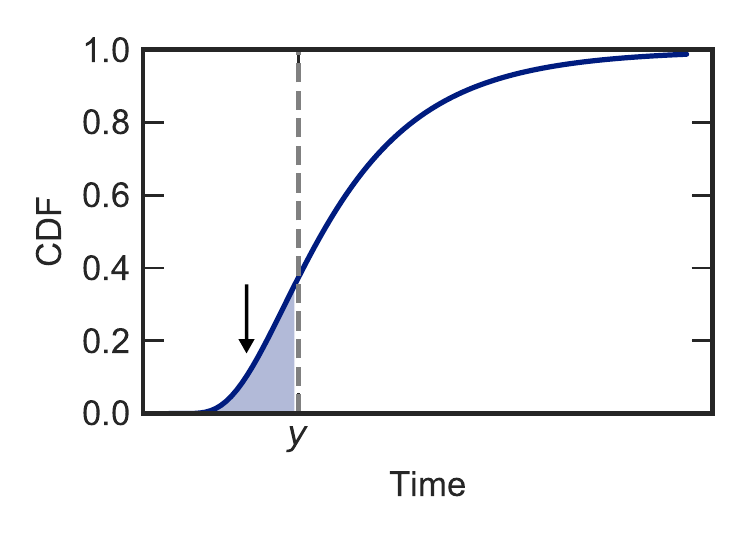}
\end{subfigure}
\begin{subfigure}[h]{0.34 \linewidth}
\caption{Interval-censored}
\label{fig:crps-c}
\includegraphics[width=0.9\linewidth]{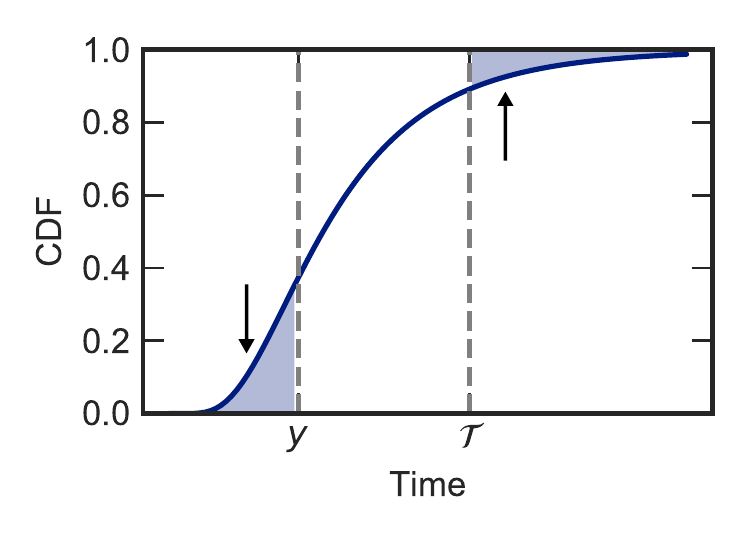}
\end{subfigure}
\caption{Graphical intuition for the Survival-CRPS scoring rule. For uncensored observations, we minimize mass before and after the observed time of event. For right-censored observations, we minimize mass before observed time of censoring. For interval-censored observations, we minimize mass before observed time of censoring, and mass after the time by which event must have occurred.}
\label{fig:crps}
\end{figure*}

A scoring rule is a measure of the quality of a probabilistic forecast. A forecast over a continuous outcome is a probability density function over all possible outcomes, $\hat{f}$ with corresponding cumulative density function $\hat{F}$. In reality, we observe some actual outcome, $y$. A scoring rule $\mathcal{S}$ takes a predicted distribution and an actual outcome, and returns a loss $\mathcal{S}(\hat{F}, y)$. It is considered a \emph{proper scoring rule} if for all possible distributions $G$,
\begin{align*} \mathbb{E}_{y\sim \hat{F}}[\mathcal{S}(\hat{F}, y)] \leq \mathbb{E}_{y\sim \hat{F}}[\mathcal{S}(G, y)],\end{align*} and \emph{strictly} proper when equality holds if and only if $\hat{F} = G$ \citep{gneiting_probabilistic_2014}. A proper scoring rule is one in which the expected score is minimized by the distribution with respect to which the expectation is taken. Intuitively, it encourages a model for being honest by predicting what it actually believes \citep{SavageElicitation}. When a proper scoring rule is employed as a loss function, it naturally rewards the model for outputting calibrated probabilities \citep{gneiting_probabilistic_2014}.

There are many commonly used proper scoring rules. Perhaps the most widely used is the logarithmic scoring rule, equivalent to the MLE objective:
\begin{align*}\mathcal{S}_{\text{MLE}}(\hat{F}, y) = -\log \hat{f}(y).\end{align*}
In the setting with censored data, we maximize the density for observed outcomes, and tail or interval mass for censored outcomes, and this is a proper scoring rule \citep{dawid_theory_2014}.
\begin{align*}
    \mathcal{S}_\text{MLE-RIGHT}(\hat{F}, (y, c)) & = -\log \big( (1-c) \hat{f}(y) \\&\quad +c (1-\hat{F}(y)) \big)\\
    \mathcal{S}_\text{MLE-INTVL}(\hat{F}, (y, c, \mathcal{T})) & = -\log \big( (1-c)\hat{f}(y) \\&\quad + c (\hat{F}(\mathcal{T}) - \hat{F}(y))\big)
\end{align*}
However, the logarithmic scoring rule is asymmetric, and harshly penalizes predictions that are wrong yet confident. 
Specifically, when the true data generating process is heavier tailed than the assumed data generating process, the training process becomes sensitive to outliers and yields more conservative (that is, less sharp) predictions as a result \citep{gebetsberger_estimation_2018}.

Another proper scoring rule for forecasts over continuous outcomes is the CRPS \citep{ProbForeCalibSharp}, defined as
\begin{align*}\mathcal{S}_{\text{CRPS}}(\hat{F}, y) &=  \int_{-\infty}^{\infty} \left(\hat{F}(z) - \mathbb{1}\{z \ge y\}\right)^2 dz \\ &= \int_{-\infty}^y \hat{F}(z)^2dz + \int_{y}^{\infty}(1-\hat{F}(z))^2dz.\end{align*}
The CRPS has been used in regression as an objective function that yields sharper predicted distributions compared to MLE, while maintaining calibration \citep{gneiting_probabilistic_2014}. Intuition for the CRPS is better understood by analyzing the latter expression and noting that the two integral terms correspond to the two shaded regions in Figure \ref{fig:crps}\subref{fig:crps-a}. The CRPS score is completely reduced to zero when the predicted distribution places all the mass on the point of true outcome, or equivalently, when the shaded region completely vanishes.

In the context of time to event predictions we propose the \emph{Survival-CRPS} which accounts for the possibility of right-censored or interval-censored data: 
\begin{align*}
    \mathcal{S}_{\text{CRPS-RIGHT}}(\hat{F}, (y, c)) 
    &= \int_{0}^{ y} \hat{F}(z)^2dz \\ &+ (1-c)\int_{ y}^\infty (1-\hat{F}(z))^2dz,\\
    \mathcal{S}_{\text{CRPS-INTVL}}(\hat{F},(y,c,\mathcal{T})) 
    &= \int_{0}^{y} \hat{F}(z)^2dz \\ & + (1-c)  \int_{y}^{\mathcal{T}} (1-\hat{F}(z))^2dz \\&+ \int_{\mathcal{T}}^\infty (1-\hat{F}(z))^2dz.
\end{align*}

Note that when $c=0$, both of the above expressions are equivalent to the original CRPS. Again, the intuition behind the Survival-CRPS is better understood by  mapping each of the integral terms to the corresponding shaded region in Figure \ref{fig:crps}\subref{fig:crps-b} and Figure \ref{fig:crps}\subref{fig:crps-c}. The Survival-CRPS behaves like the original CRPS when the time of event is uncensored. For censored outcomes, it penalizes the predicted mass that occurs before the time of censoring and, if interval censored, also the mass after time by which the event must have occurred.

Both variants of the Survival-CRPS are proper scoring rules. They are special cases of the threshold weighted CRPS \citep{Ranjan_2011}, where the weighting function is an indicator over the uncensored regions.

\subsection{\textsc{Evaluation by Sharpness subject to Calibration}}

\emph{Calibration} assesses how well forecasted event probabilities match up to observed event probabilities. It is crucial in development of useful predictive models, especially for clinical decision-making. 
In binary prediction tasks without censoring, the Hosmer-Lemeshow test statistic \citep{hosmer_lemeshow} is commonly used to assess goodness-of-fit by comparing observed versus predicted event probabilities at quantiles of predicted probabilities. Extensions to account for censoring have been proposed \citep{Gronnesby1996, dagostnio, demler_tests_2015}, but these methods apply only to binary predictions for a particular time frame (for example, 1-year risks of mortality).

There is no widely accepted method for evaluating how well calibrated a set of entire prediction distributions is in the time to event setting. D-calibration has been recently proposed as a method for holistic evaluation \citep{andres_novel_2018}, but relies on handling censored observations by assuming the true times to event are uniformly distributed past the times of censoring in the predicted distributions. When censored observations far outnumber the uncensored observations, this can lead to overly optimistic assessments of calibration. Another option is to evaluate observed event times on the cumulative density scale of predicted distributions, using a Kaplan-Meier estimate to account for censoring \citep{harrell_rms}. Again, this method has limitations in the heavily censored setting, as the quantiles in the tail of predicted cumulative densities have few uncensored observations, and will rarely yield well calibrated values. 

We instead employ the following method to measure calibration. We compare predicted cumulative densities against observed event frequencies, evaluated at quantiles of predicted cumulative density. Right-censored observations are removed from consideration in quantiles that correspond to times after their points of censoring. Interval-censored observations are similarly removed from consideration in quantiles that correspond to times after censoring, but are additionally re-introduced in quantiles that correspond to times past the time by which the event must have occurred (in the mortality prediction task, this corresponds to 120 years of age). In this work we assess resulting calibration curves qualitatively by graphing them, and quantitatively by comparing the slopes of the corresponding lines of best fit (ideally 1).

Subject to calibration, we strive for prediction distributions that are  \emph{sharp} (i.e, concentrated). There are several metrics that could be used for measuring sharpness, such as variance or entropy. In the context of time to event predictions, holding two distributions with vastly different means to the same standard of variance or entropy would be unfair (for example, we would want lower variance for a prediction distribution with a mean of a day, compared to a mean of a year). Instead, we use the coefficient of variation (CoV) as a reasonable measure of sharpness. The CoV is defined as the ratio of one standard deviation to the mean,
$\text{CoV}(\hat{F}) = \frac{\sqrt{\text{Var}[\hat{F}]}}{\mathbb{E}[\hat{F}]}.$

\subsection{\textsc{Survival-AUPRC: holistic time to event metric}}\label{survauprc}

\begin{figure*}[t]
\centering
\begin{subfigure}[h]{0.35 \linewidth}
\caption{Analog of recall for precision of 0.9.}
\label{fig:auprc-a}
\includegraphics[width=\linewidth]{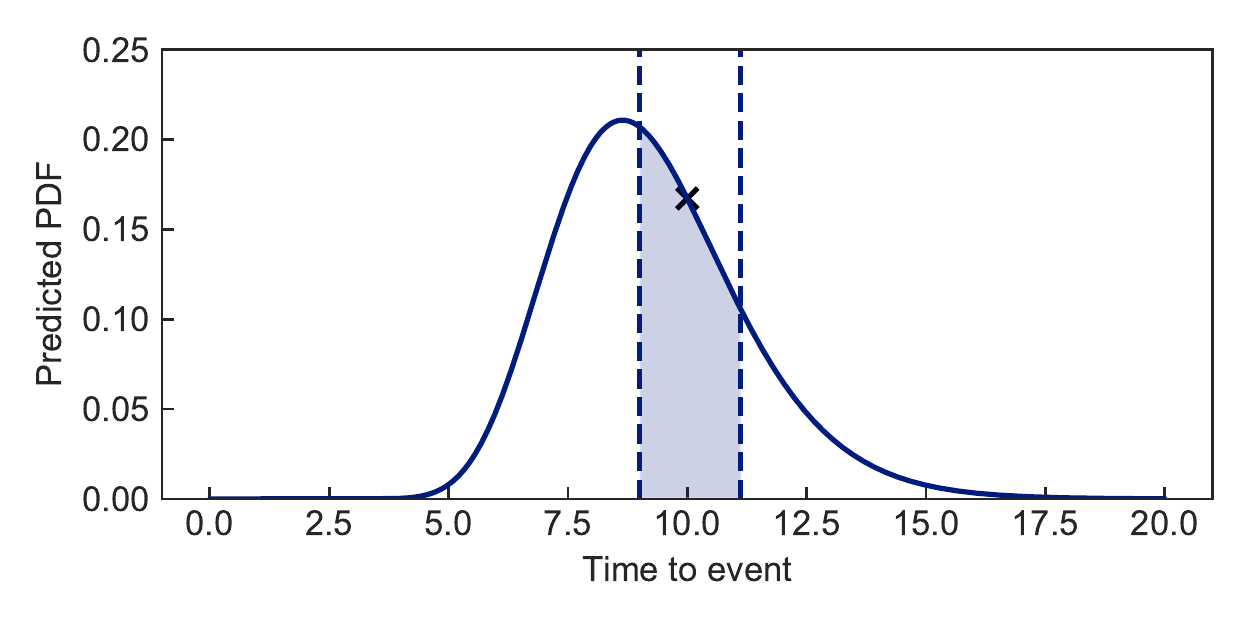}
\end{subfigure}
\begin{subfigure}[h]{0.35 \linewidth}
\caption{Corresponding precision-recall plot.}
\label{fig:auprc-b}
\includegraphics[width=\linewidth]{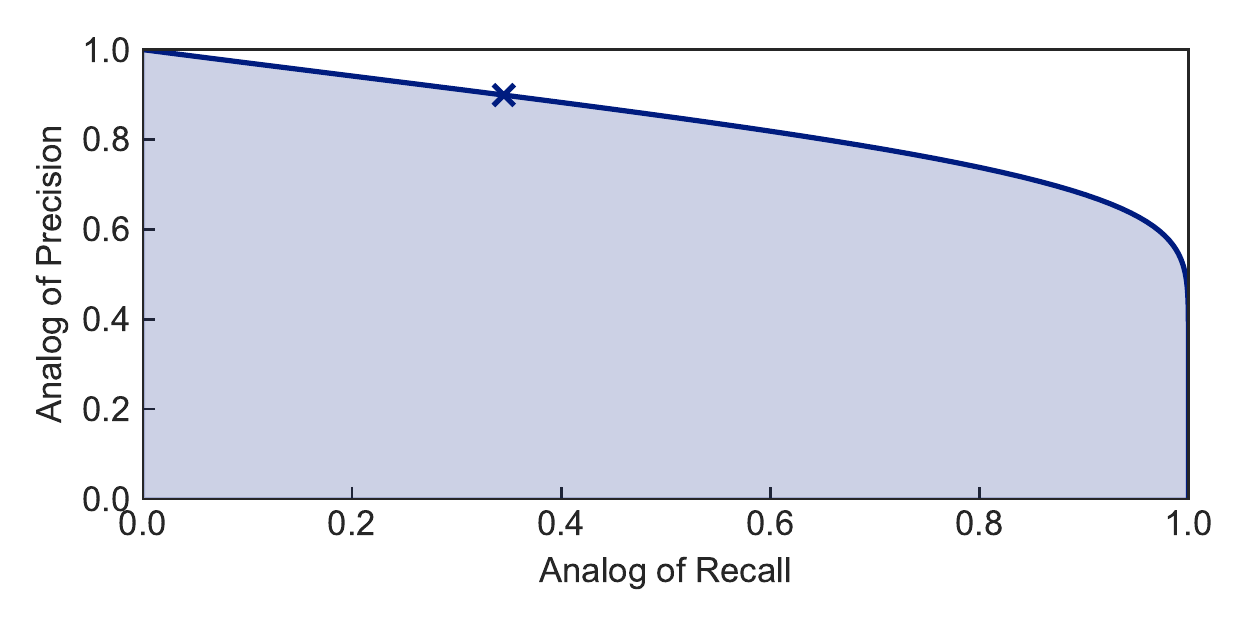}
\end{subfigure}
\caption{Graphical intuition for the Survival-AUPRC metric for uncensored observations. We compute the mass that lies within each interval of precision around the true outcome to compute the precision-recall curve.}
\label{fig:auprc}
\end{figure*}

Since sharpness is only a function of the predicted distributions, a measure of sharpness is only meaningful if the model is sufficiently calibrated. We now propose a metric that measures how concentrated the mass of the prediction distribution is around the true outcome, robust to miscalibration. The idea is similar to the area under a precision-recall curve, except here it is with respect to only one predicted distribution and one outcome. We first consider the uncensored case. As an analog to precision, we consider intervals relative to the true time of event, defined by ratios. For example, a region of precision 0.9 around an event that occurs at time $y$ is the interval $[0.9y, y/0.9]$. Corresponding to this region of precision, the analogy to recall is the mass assigned by the predicted distribution over this interval, $\hat{F}(y/0.9) - \hat{F}(0.9y)$. By exploring the full range of precision from 0 to 1, we obtain the \emph{Survival Precision Recall Curve}. The area under this curve measures how quickly predicted mass concentrates around the true outcome as we expand the precision window. 
\begin{align*}
    \text{Surv-AUPRC}_{\text{UNCENS}}(\hat{F}, y) &= \int_0^1 (\hat{F}(y/t) - \hat{F}(yt))dt
\end{align*}
The highest possible score is 1, when the predicted distribution is a Dirac $\delta$ function centered over the time of outcome. The lowest possible score is 0, when the predicted distribution is infinitely dispersed. The mean of all Survival-AUPRC scores across examples provides an overall measure of the quality of the predictions. 

The aforementioned metric only applies when the event outcome is uncensored. In the case of censored observations, we use the same analogy but with the right end of precision intervals defined with respect to the time by which the event must have occurred in the interval-censored case, or infinity in the right-censored case.
\begin{align*}
\text{Surv-AUPRC}_{\text{RIGHT}}(\hat{F}, y) &= \int_0^1 (1 - \hat{F}(yt)) dt\\
\text{Surv-AUPRC}_{\text{INTVL}}(\hat{F}, y, \mathcal{T}) &=\int_0^1 (\hat{F}(\mathcal{T}/t) - \hat{F}(yt)) dt
\end{align*}



\begin{figure*}[t]
    \centering
    \begin{subfigure}[b]{0.49 \linewidth}
        \caption{Dead (uncensored) patients.}
        \includegraphics[width=\linewidth]{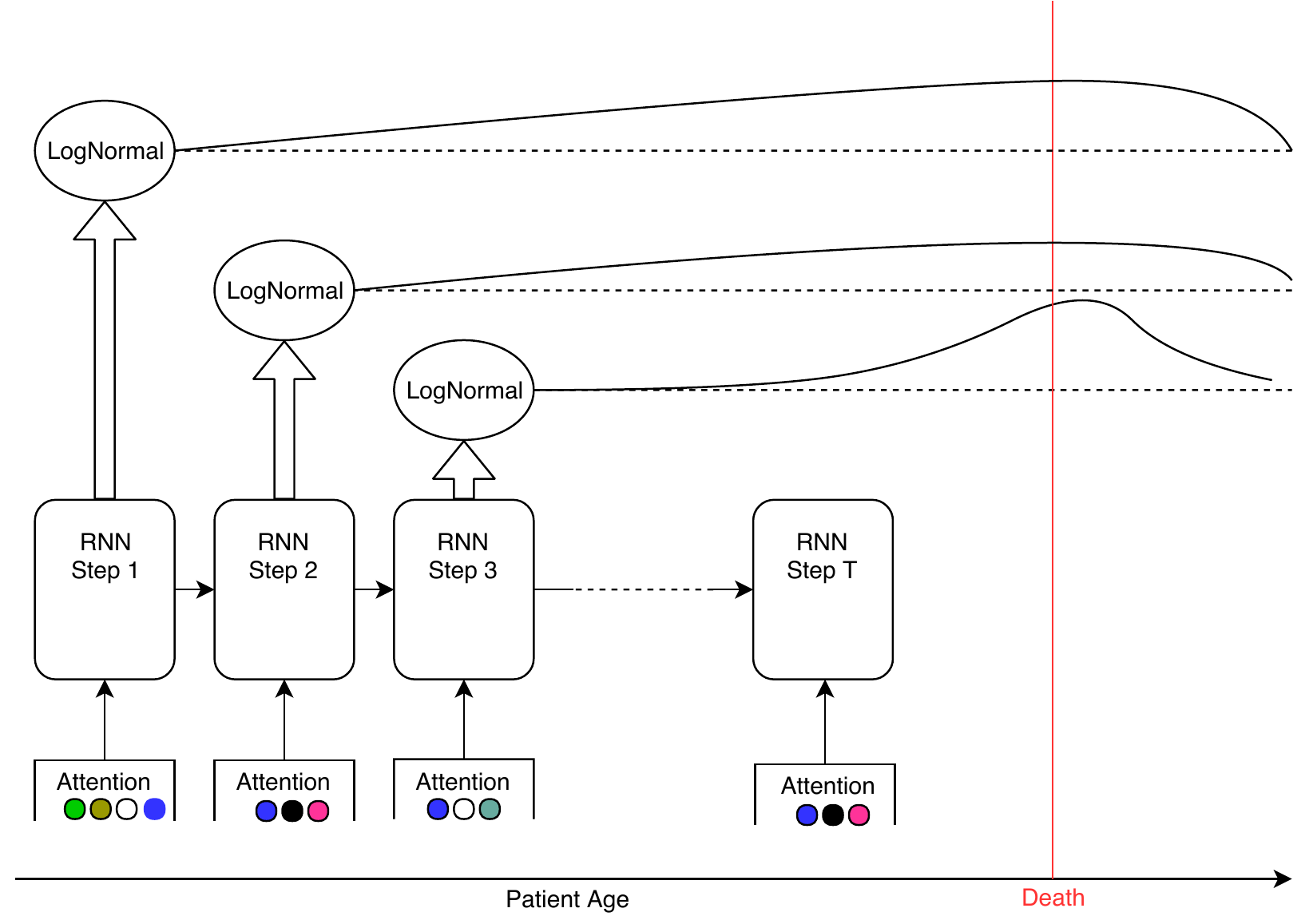}
    \end{subfigure}
    \\
    \begin{subfigure}[b]{0.49 \linewidth}
        \caption{Alive (right-censored) patients.}
        \includegraphics[width=\linewidth]{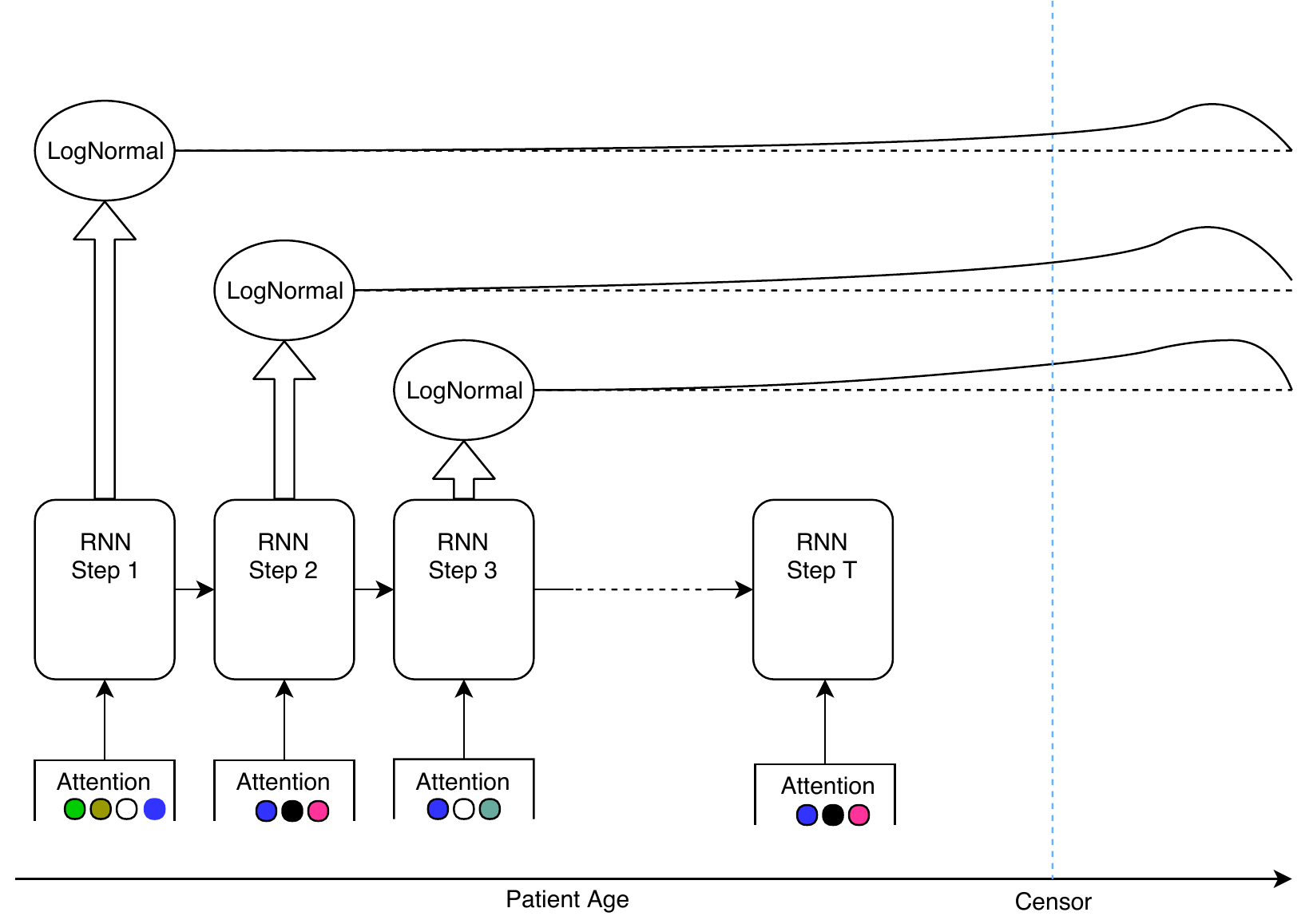}
    \end{subfigure}
    \begin{subfigure}[b]{0.49 \linewidth}
        \caption{Alive (interval-censored) patients.}
        \includegraphics[width=\linewidth]{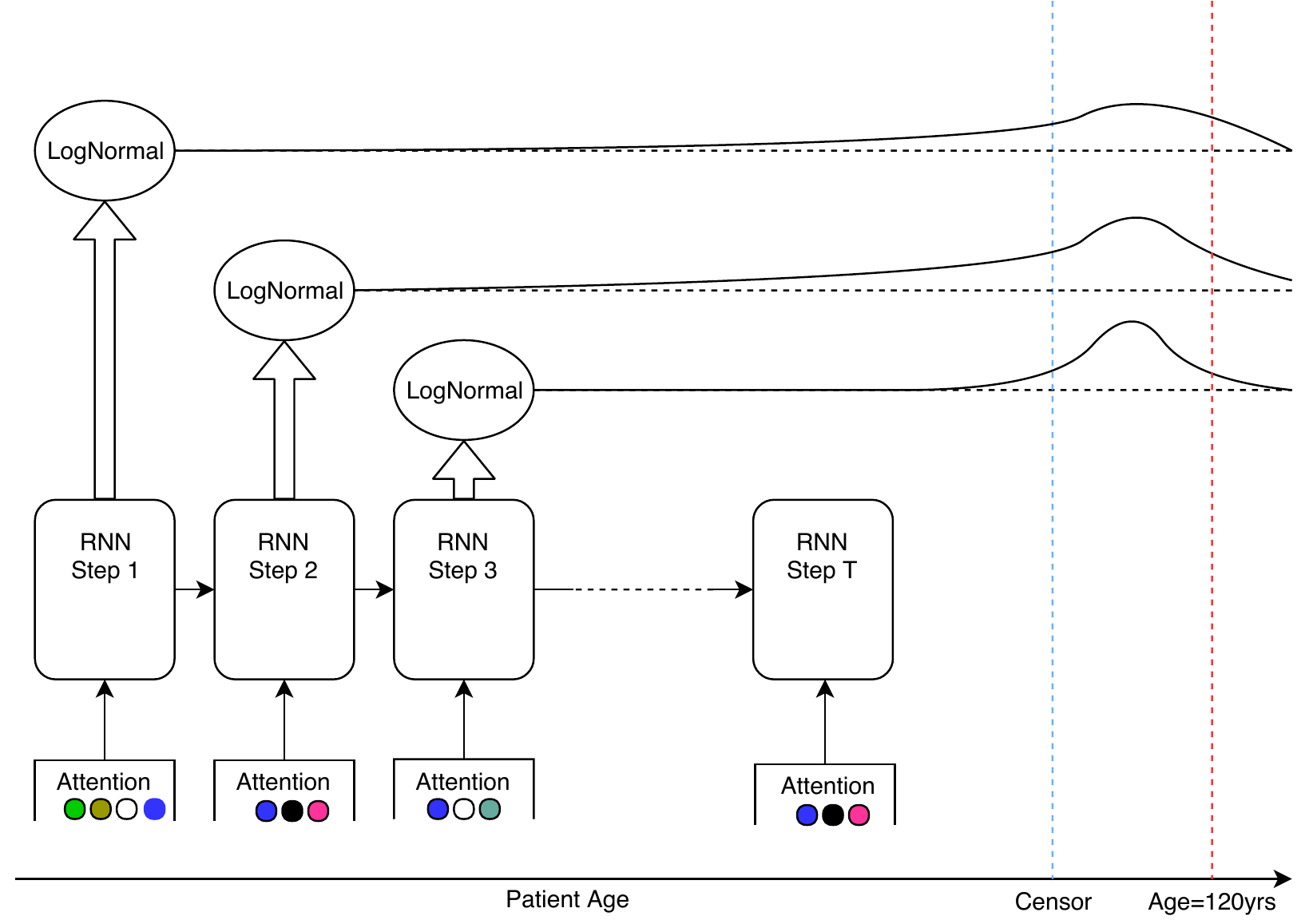}
    \end{subfigure}
    \caption{At each time step, we predict parameters $\mu, \sigma^2$ of a log-normal distribution, minimizing a proper scoring rule. While (a) shows how to handle dead (uncensored) patients, (b) and (c) compare right and interval censorship. Leveraging knowledge about the world (e.g. mortality occurs by 120 yrs), (c) shows that interval censorship produces a sharper distribution.}
    \label{fig:rnn_model}
\end{figure*}

\subsection{\textsc{Implementation tips and choices}}\label{recommendations}

Common parametric distributions over time to event used in traditional survival analysis models include the Weibull, Log-Normal, Log-Logistic, and Gamma. In order to be sufficiently expressive in model space, we seek distributions with at least two parameters. We recommend the Log-Normal distribution because other distributions either involve the Gamma function in their density, or involve the pattern $(y/p_1)^{p_2}$, where $p_1$ and $p_2$ are parameters output from the neural network. We found these patterns to be highly sensitive to the inputs and to suffer from numerical instability issues.

For the Log-Normal distribution, a closed form expression for the CRPS is well-known \citep{Baran_2015}. However, a closed form expression for the Survival-CRPS does not exist. We perform a change of variable to express the integral terms as finite integrals, and numerically approximate with the trapezoid rule. When training, we then back-propagate through the trapezoidal approximation. Details are given in Appendix \hyperref[sec:appendix-B]{B} and \hyperref[sec:appendix-C]{C}. Separately, we note that the approximation formulas are themselves proper scoring rules, as they are just weighted sums of Brier scores. Closed form expressions for the log-normal Survival-AUPRC are also given in Appendix \hyperref[sec:appendix-D]{D}, \hyperref[sec:appendix-E]{E}, and \hyperref[sec:appendix-F]{F}.

\section{\textsc{Experiments}}\label{experiments}

We assume a dataset of longitudinal records indexed by $i$, $\left(\{ (x^{(i)}_t, a^{(i)}_t) \}_{t=1}^{T^{(i)}}, d^{(i)}, c^{(i)}\right)$, where $t \in \{1 \hdots T^{(i)}\}$ denotes the interaction number of this patient with the health record, $x^{(i)}_t \in \mathbb{R}^D$ is the set of features corresponding to the $t$-th interaction, $a^{(i)}_t \in \mathbb{R}_+$ is age at time $t$, $d^{(i)} \in \mathbb{R}_+$ is the age of death or age of last known (alive) encounter, and $c^{(i)} \in \{0,1\}$ is a censoring indicator where $c^{(i)} = 0$ means the age of death is $d^{(i)}$, and $c^{(i)} = 1$ means the age of death is at least $d^{(i)}$. For each $x^{(i)}_t$ we define the quantity $y^{(i)}_t = d^{(i)} - a^{(i)}_t$ which represents the corresponding time to event or time to censoring. For interval censoring we assume a maximum lifespan of 120 years.

We run experiments for the mortality prediction task to evaluate four different training objectives: Maximum Likelihood $\mathcal{S}_\text{MLE-RIGHT}$ and $\mathcal{S}_\text{MLE-INTVL}$, and our Survival-CRPS based loss $\mathcal{S}_\text{CRPS-RIGHT}$ and $\mathcal{S}_\text{CRPS-INTVL}$. 

\subsection{\textsc{RNN with STARR Data Warehouse}}
We use electronic health records from the STARR Data Warehouse (previously known as STRIDE) for training and evaluation \citep{Lowe2009STRIDEPlatform}. The Warehouse contains de-identified data for over 3 million patients (about 2.6\% having a recorded date of death), spanning approximately 27 years. The data we use on patients include diagnostic codes, medication order codes, lab test order codes, encounter type codes, and demographics (age and gender).

The structure of this dataset motivates our model selection; because many patients in this dataset have records spanning multiple days and visits, we use a recurrent neural network (RNN) on this particular task. We assign timesteps to each day that a patient has recorded data. The set of 3 million patients, correspond to 51 million overall timesteps, and was randomly split in the ratio 8:1:1 into train, validation and test splits. 
Formally, our model $\textrm{RNN}$ is parameterized by $\theta$, denoted $\textrm{RNN}_\theta$, that takes as input a sequence of features to predict parameters of a parametric probability distribution $\hat{F}$ over time to death at each timestep (Figure \ref{fig:rnn_model}). The network depends only on data from the current and previous timesteps, and not the future. The approach here is similar to the recently proposed Weibull Time to Event RNN \citep{martinsson_model_2016}, though we generalize to any choice of noise distribution. The distributions that are output in each timestep are used to construct an overall loss,
\begin{align*}
\mathcal{L}_{\text{RIGHT}} & = \sum_{i=1}^N \sum_{t=1}^{T^{(i)}} \mathcal{S}_{\text{RIGHT}}\left(\hat{F}_{\text{RNN}_{\theta} \left\{ x_{1:t}^{(i)} \right\}},  \big(y_t^{(i)}, c^{(i)}\big)\right)\\
\mathcal{L}_{\text{INTVL}} & = \sum_{i=1}^N \sum_{t=1}^{T^{(i)}} \mathcal{S}_{\text{INTVL}}\left(\hat{F}_{\text{RNN}_{\theta} \left\{ x_{1:t}^{(i)} \right\}},  \big(y_t^{(i)}, c^{(i)}, \mathcal{T}^{(i)}_t\big)\right),
\end{align*}
where $N$ is the total number of patients in the training set, $T^{(i)}$ is the sequence length for patient $i$, and $\hat{F}_{\text{RNN}_{\theta}}$ denotes the distribution parameterized by the output of the RNN. 

We leverage a combination of real-valued (e.g. age of patient) and discrete (e.g., ICD codes) data. Discrete data is embedded into a trainable real-valued 126-dimensional vector space. The vectors of ICD codes at a single timestep are combined into a weighted mean by a soft self-attention mechanism. All real-valued inputs are appended to the averaged embedding vector. We also provide the real-valued features to every layer by appending them to the output of previous layer. The input vector feeds into a fully connected layer, followed by multiple recurrent layers. We use the Swish activation function \citep{swish} and layer normalization \citep{layernorm} at every layer. Recurrent layers are defined using GRU units \citep{gru} with layer normalization inside. After the set of recurrent layers, the network has multiple branches, one per parameter of the survival distribution (for the lognormal, $\mu$ and $\sigma^2$). The final layer in each branch has scalar output, optionally enforced positive with the softplus function, $\text{Softplus}(z) = \log(1+\exp(z))$. We use Bernoulli dropout \citep{dropout} at all fully connected layers, and Variational RNN dropout \citep{variational-rnn} in the recurrent layers, with a dropout probability of 0.5. Optimization is performed using the Adam optimizer \citep{kingma_ba}, with a fixed learning rate of 1e-3.

\subsection{\textsc{FCN with MIMIC-III}}

We leverage the publicly available MIMIC-III dataset \citep{mimic}. This dataset differed from the STARR EHRs in that patients were often admitted only once and whose visits did not exceed a day. To match this structure of this dataset, we build a 4 layer feed forward neural network that takes in $51015$ hospital admissions in the dataset ($70.1\%$ censored) and makes predictions at the time of discharge. We removed admissions where the patient's age was obfuscated or where the patient's discharge time occurred after their recorded date of death. As features, we used demographics (age as real-valued, and gender as binary) and embedded diagnostic codes into a 128-dimensional space.

\subsection{\textsc{Results}}

\begin{table*}[!]
  \centering
  \caption{Metrics measuring sharpness and calibration for models trained on the right-censored and interval-censored variants of the Maximum Likelihood and Survival-CRPS objectives. Confidence intervals are generated by bootstrap resampling of the test set.}
  \label{table-sharpness}
  \vspace{0.1cm}
  \small
  STARR Dataset (97.4\% censoring) \\
  \vspace{0.1cm}
  \begin{tabular}{lcccc}
    \toprule
    Metric & MLE-RIGHT & MLE-INTVL & CRPS-RIGHT & CRPS-INTVL\\ \midrule
    Calibration slope & 1.125 $\pm$ 3e-4 & 1.139 $\pm$ 3e-4 & 1.003 $\pm$ 3e-4 & 0.959 $\pm$ 5e-4 \\ 
    Mean coefficient of variation & \textcolor{BrickRed}{18.42 $\pm$ 5e-3} & 0.911 $\pm$ 4e-4 & 0.332 $\pm$ 1e-4 & \bf{0.301 $\pm$ 1e-4} \\
    Mean prob of survival to age 120 yrs & \textcolor{BrickRed}{0.754 $\pm$ 2e-5} & 0.045 $\pm$ 3e-5 & 0.015 $\pm$ 3e-5 & \bf{0.005 $\pm$ 1e-6} \\
    Dead: mean Surv-AUPRC (uncen) & 0.233 $\pm$ 2e-4 & 0.319 $\pm$ 3e-4 & 0.343 $\pm$ 4e-4 & \bf{0.366 $\pm$ 4e-4} \\
    Alive: mean Surv-AUPRC (intvl-cen) & 0.407 $\pm$ 6e-5 & 0.963 $\pm$ 2e-5 & \bf{0.977 $\pm$ 3e-5} & 0.976 $\pm$ 3e-5 \\
    \bottomrule
  \end{tabular}
   \vspace{0.2cm}
  \small 
  \\MIMIC-III Dataset (70.1\% censoring) \\
  \vspace{0.1cm}
  \begin{tabular}{lcccc}
    \toprule
    Metric & MLE-RIGHT & MLE-INTVL & CRPS-RIGHT & CRPS-INTVL\\ \midrule
    Calibration slope & 0.945 $\pm$ 9e-3 & 0.933 $\pm$ 1e-2 & 0.951 $\pm$ 1e-2 & 0.938 $\pm$ 1e-2\\ 
    Mean coefficient of variation & 2.218 $\pm$ 0.011 & 1.763 $\pm$ 0.006 & 1.797 $\pm$ 0.014 & \bf{1.647 $\pm$ 0.012} \\
    Mean prob of survival to age 120 yrs & {0.012 $\pm$ 4e-4} & 0.007 $\pm$ 2e-4 & 0.001 $\pm$ 2e-4 & \bf{0.001 $\pm$ 3e-5} \\
    Dead: mean Surv-AUPRC (uncen) & 0.329 $\pm$ 2e-3 & 0.338 $\pm$ 4e-3 & 0.342 $\pm$ 3e-3 & \bf{0.348 $\pm$ 4e-3} \\
    Alive: mean Surv-AUPRC (intvl-cen) & 0.993 $\pm$ 2e-4 & 0.999 $\pm$ 6e-5 & 1.000 $\pm$ 2e-4 & \bf{1.000 $\pm$ 1e-5} \\
    \bottomrule
  \end{tabular}
\end{table*}

\begin{figure*}[!t]
  \centering
    \small
    STARR Dataset \\
  \includegraphics[width=0.7\linewidth]{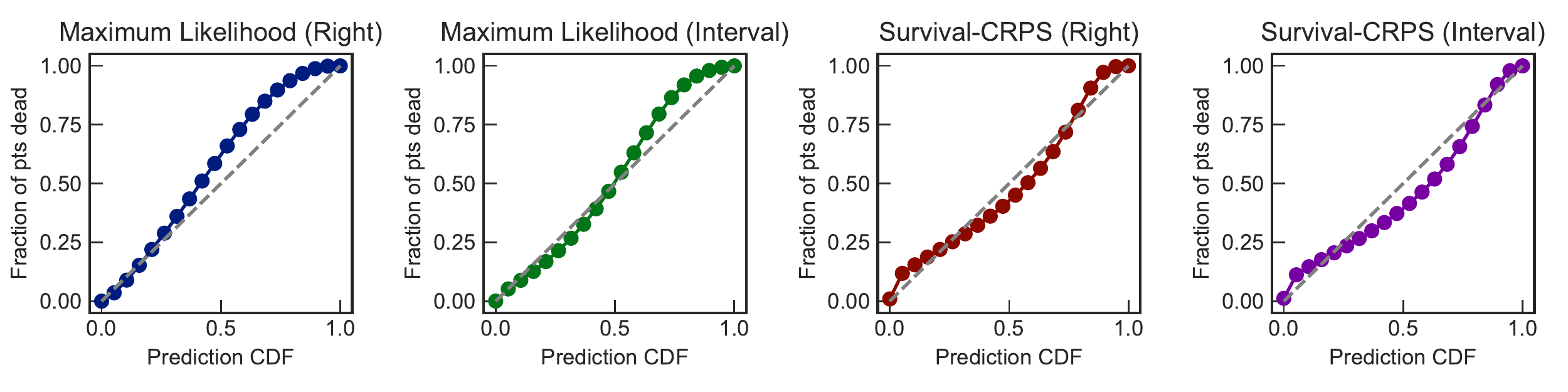}
  \centering
  \vspace{0.2cm}
  \\MIMIC-III Dataset \\
  \includegraphics[width=0.7\linewidth]{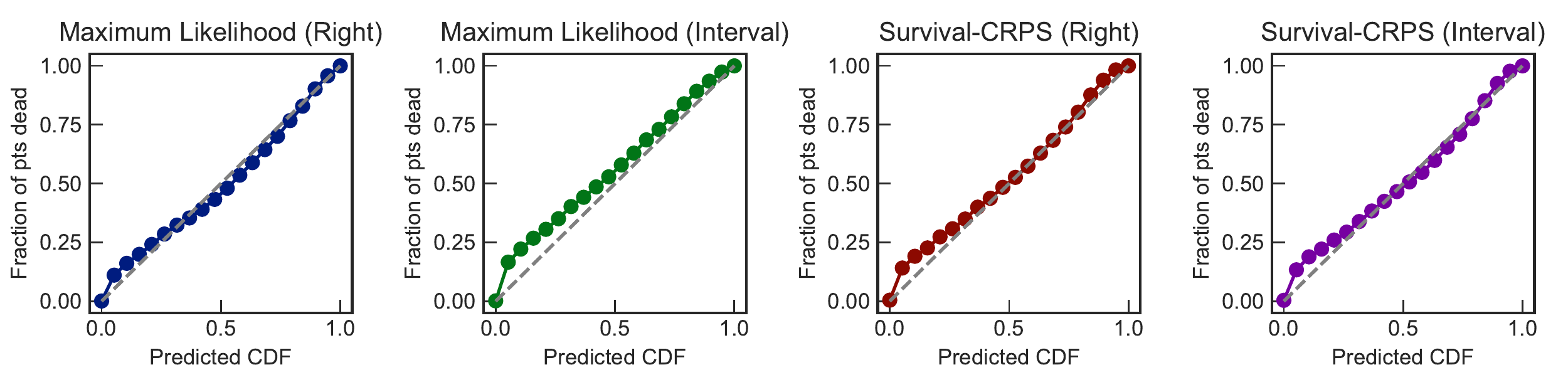}
  \caption{Calibration plots for each of the models. We compare predicted cumulative densities against observed event frequencies, evaluated at quantiles of predicted cumulative density. Right-censored observations are removed from consideration in quantiles past times of censoring, interval-censored observations are additionally re-introduced in quantiles corresponding to times past 120 years.}
  \label{fig-calib}
\end{figure*}

The results are presented in Table \ref{table-sharpness}. Both the coefficient of variation and the Survival-AUPRC metrics suggest that the Survival-CRPS with interval censoring yields the sharpest prediction distributions. Inspecting the probability past 120 years of age for the STARR dataset shows that a naively trained prediction model with MLE can assign more than 75\% of the mass to unreasonable regions, which is highly undesirable for the purpose of prediction. We note that this behavior is largely due to low prevalence of uncensored examples, which is typical in real-world EHR data sets. As a result, the loss for the censored examples, which can be minimized by pushing mass as far away to the right as possible, dominates the small number of uncensored examples. While the benefits of Survival-CRPS are most pronounced in low prevalence datasets (STARR), they show benefit with moderate prevalence as well (MIMIC-III).

\begin{figure}[h]
  \centering
  \begin{subfigure}[b]{0.8 \linewidth}
    \includegraphics[width=\linewidth]{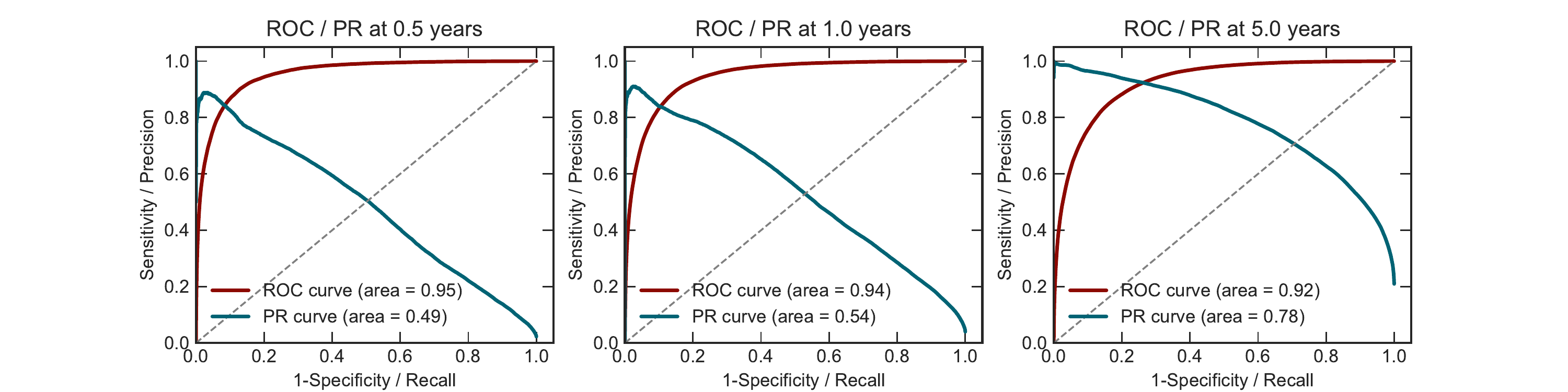}
  \end{subfigure}
  \begin{subfigure}[b]{0.8 \linewidth}
    \includegraphics[width=\linewidth]{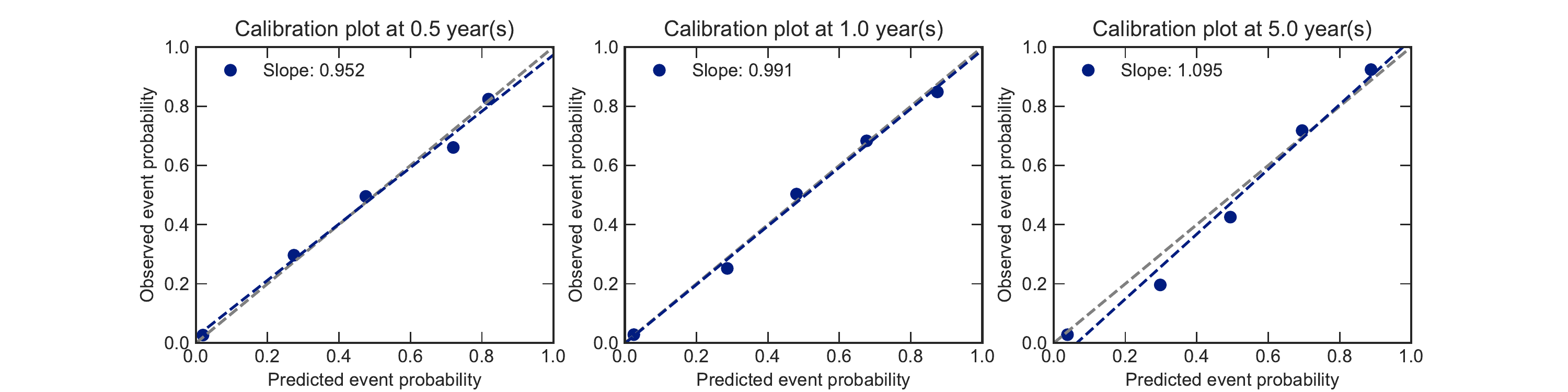}
  \end{subfigure}
  \caption{Discrimination and calibration of predictions from the interval-censored Survival-CRPS model, evaluated as predictions for a dichotomous outcome at 6 months, 1 year, and 5 years.}
  \label{fig-dichotomous}
\end{figure}

Since we employ proper scoring rules, the predictions tend to be well calibrated (Figure \ref{fig-calib}). By predicting an entire distribution over time to an indivdual's mortality, the same model can be used to make classification predictions at various time points, highlighting the flexibility of our approach. When evaluated at 6 month, 1 year, and 5 year probabilistic predictions of mortality, our model remains well-calibrated with high discriminative ability (Appendix \hyperref[sec:appendix-G]{G}, Figure \ref{fig-dichotomous}).

\def\githuburl{http://github.com/stanfordmlgroup/cdr-mimic}

Source code of our implementation is published  \footnote{\href{\githuburl}{\tiny\texttt{\githuburl{}}}}. 
 
\section{\textsc{Related Work}}

Recent work has demonstrated potential to significantly improve patient care by making predictions with deep learning models on EHR data \citep{avati_improving_2017, rajkomar_scalable_2018}, but these works have been limited to treating the task as binary classification over a fixed time frame. Predicting survival curves instead of dichotomous outcomes has been explored \citep{yu_learning_2011, lee_deephit:_2018}, but these works predict over a discrete set of times. Work in \citep{NIPS2017_6827} also predicts full survival curves specific to a patient, but the use of Gaussian Processes makes it difficult to scale to datasets with millions of patients. Deep survival analysis \citep{ranganath_deep_2016} has been proposed, but is limited to a fixed shape Weibull (bypassing the concerns we raised about stability, but limited in expressivity). The work by \cite{pmlr-v68-yang17a} is similar to ours in terms of using log-normal noise distribution, but is limited to MLE training. DeepSurv \citep{katzman_deepsurv:_2018} uses a Cox proportional hazards model, which similarly makes a set of inflexible assumptions. The WTTE-RNN \citep{martinsson_model_2016} model is also limited to a Weibull distribution. All aforementioned models have only been optimized by MLE, instead of more robust proper scoring rules. The CRPS scoring rule has been used with neural networks in \citep{rasp_lerch}. The work in \cite{Miscouridou2018DeepSA} predicts survival curves (both non-parametric, and flexible flow based parametric curves) while also handling missing covariates. Another recent work \cite{chapfuwa2018adversarial} uses adversarial training for survival prediction. It has been shown that modern neural networks can be miscalibrated, and the work by \cite{DBLP:journals/corr/GuoPSW17} and \cite{kuleshov_accurate_2018} suggest ways to improve calibration.

\section{\textsc{Conclusion}}

Better survival prediction models can be built by exploring objectives beyond MLE and evaluation metrics that assess the holistic quality of predicted distributions, instead of point estimates. We introduced the Survival-CRPS objective, motivated by the fact that the CRPS scoring rule is known to yield sharp prediction distributions while maintaining calibration. There are perhaps others scoring rules that work better, leaving avenues for future work. To evaluate, we introduced the Survival-AUPRC metric, which captures the degree to which a prediction distribution concentrates around the observed time of event. We demonstrate large-scale survival prediction by using a deep models employing a log-normal parameterization. The impact of having meaningfully accurate survival models is tremendous, especially in healthcare. We hope our work will be useful to those looking to build and deploy such models. 

\subsubsection*{\textsc{Acknowledgments}}

We thank Baran Sandor, Sebastian Lerch, Alejandro Schuler, Jeremy Irvin, and Russell Greiner for valuable feedback.


\allowdisplaybreaks
\clearpage
\def\FN{\ensuremath F_{\mathcal{N}(\mu,\sigma^2)}}
\def\fN{\ensuremath f_{\mathcal{N}(\mu,\sigma^2)}}

\onecolumn
\section*{\textsc{Appendix}}

\subsection*{\textsc{A. Integral Identities}}
\label{sec:appendix-A}

Let $\Phi_{\mu,\sigma^2}(z)$ be the CDF of a Gaussian distribution with mean $\mu$ and variance $\sigma^2$. Hence $\Phi_{\mu,\sigma^2}(\log z)$ is the CDF of a log-normal distribution with mean $\mu$ and variance $\sigma^2$. For some integer $K$ (typically 32 in our experiments), we define $I$ to be the following integral, approximated by the trapezoidal rule:
\begin{align*}
    I_{\mu,\sigma^2}(y,g) &= \int_0^y \Phi_{\mu,\sigma^2}(\log z)^2g(z)dz \\
    &\approx \sum_{k=0}^{K-1}\frac{1}{2}\left[\Phi_{\mu,\sigma^2}\left(\log z_{k+1}\right)^2g\left(z_{k+1}\right) + \Phi_{\mu,\sigma^2}\left(\log z_k\right)^2g\left(z_k\right)\right](z_{k+1} - z_k)
\end{align*}
where $0 = z_0 < z_1 < ... < z_K = y$ and $g$ is a function. We further define
\begin{align*}
    I_{\mu,\sigma^2}^{+}(y) &= I_{\mu,\sigma^2}(y, z\mapsto z), \\
    I_{\mu,\sigma^2}^{-}(y) &= I_{-\mu,\sigma^2}(1/y, z\mapsto 1/z^2).
\end{align*}
\subsection*{\textsc{B. Survival-CRPS for log-normal (right-censored)}}
\label{sec:appendix-B}

For a general continuous prediction distribution $F$, with actual time to outcome $y \in \mathbb{R}_+$, and censoring indicator $c$, we generalize the CRPS to the Right Censored Survival CRPS score as:
\begin{align*}
    \mathcal{S}_{\text{CRPS-RIGHT}}(F, (y, c)) &= \int_{-\infty}^{\infty} (F(z)\mathbb{1}\{z \le \log{y} \cup c = 0\} - \mathbb{1}\{z \ge \log{y} \cap c = 0\})^2 dz \\
    &= \int_{-\infty}^{\tilde{y}} F(z)^2dz + (1-c)\int_{\tilde{y}}^\infty (F(z)-1)^2dz.
\end{align*}
In the above expression $F$ would generally be in the family of continuous distributions over the entire real line (eg. Gaussian). Alternately, one could also use a family of distributions over the positive reals (e.g log-normal), in which case the Survival CRPS becomes:
\begin{align*}
    \mathcal{S}_{\text{CRPS-RIGHT}}(F, (y, c)) &= \int_{0}^{\infty} (F(z)\mathbb{1}\{z \le y \cup c = 0\} - \mathbb{1}\{z \ge y \cap c = 0\})^2 dz \\
    &= \int_{0}^{ y} F(z)^2dz + (1-c)\int_{ y}^\infty (F(z)-1)^2dz.
\end{align*}
For the case of $F$ being log-normal, the expression becomes
\begin{align*}
    \mathcal{S}_{\text{CRPS-RIGHT}}(F_{\text{LN}(\mu, \sigma^2)}, (y, c)) &= \int_{0}^{y} \Phi_{\mu,\sigma^2}(\log z)^2dz + (1-c) \int_{y}^{\infty} (1-\Phi_{\mu,\sigma^2}(\log z))^2dz \\
    &= \int_{0}^{y} \Phi_{\mu,\sigma^2}(\log z)^2dz + (1-c) \int_{y}^{\infty} \Phi_{-\mu,\sigma^2}(-\log z)^2dz \\
    &= \int_{0}^{y} \Phi_{\mu,\sigma^2}(\log z)^2dz + (1-c) \int_{0}^{1/y} \Phi_{-\mu,\sigma^2}(\log z)^2(1/z)^2dz
    \\
    &= I_{\mu,\sigma^2}^{+}(y) + (1-c)I_{\mu,\sigma^2}^{-}(y).
\end{align*}

\subsection*{\textsc{C. Survival-CRPS for log-normal (interval-censored)}}
\label{sec:appendix-C}

We further extend the Right Censored Survival CRPS to the case of interval censoring. This is particularly useful for all-cause mortality prediction where we assume a particular event must occur by time $\mathcal{T}$. Using the same notations as before, the Interval Censored Survival CRPS is:
\begin{align*}
    \mathcal{S}_{\text{CRPS-INTVL}}(F,(y,c,\mathcal{T})) &= \int_{0}^{\infty} (F(z)\mathbb{1}\{\{z \le y \cup c = 0\} \cup z \ge \mathcal{T} \} - \mathbb{1}\{\{z \ge y \cap c = 0\} \cup z \ge \mathcal{T}\})^2 dz \\
    &= \int_{0}^{y} F(z)^2dz + (1-c)\int_{y}^{\mathcal{T}} (F(z)-1)^2dz + \int_{\mathcal{T}}^\infty (F(z)-1)^2dz.
\end{align*}

For the case of $F$ being log-normal, the expression becomes
\begin{align*}
    \mathcal{S}_{\text{CRPS-INTVL}}(F_{\text{LN}(\mu, \sigma^2)}, (y, c, \mathcal{T})) &= \int_{0}^{y} \Phi_{\mu,\sigma^2}(\log z)^2dz + (1-c) \int_{y}^{\mathcal{T}} (1-\Phi_{\mu,\sigma^2}(\log z))^2dz \\ &\qquad + \int_{\mathcal{T}}^{\infty} (1-\Phi_{\mu,\sigma^2}(\log z))^2dz \\
    &= \int_{0}^{y} \Phi_{\mu,\sigma^2}(\log z)^2dz + (1-c) \int_{1/\mathcal{T}}^{1/y}  \Phi_{-\mu,\sigma^2}(\log z)^2(1/z)^2dz \\ &\qquad + \int_{0}^{1/\mathcal{T}}  \Phi_{-\mu,\sigma^2}(\log z)^2(1/z)^2dz\\
    &= I_{\mu,\sigma^2}^{+}(y) + I_{\mu,\sigma^2}^{-}(\mathcal{T})+ (1-c)\bigg[I_{\mu,\sigma^2}^{-}(y) - I_{\mu,\sigma^2}^{-}(\mathcal{T}) \bigg].
\end{align*}

\subsection*{\textsc{D. Survival-AUPRC for log-normal (interval-censored)}}
\label{sec:appendix-D}

We start with the most general case (interval censoring). For a general continuous prediction distribution $F$ with an interval outcome $[L, U]$, we define the Survival-AUPRC as 
\begin{align*}
    \text{Survival-AUPRC}(F, L, U) &= \int_0^1\left[F(U/t) - F(Lt)\right]dt.
\end{align*}
Specifically for the case of log-normal, where $\phi$ and $\Phi$ are PDF and CDF of $\mathcal{N}(0,1)$ respectively, and $\tilde{L} = \log L$ and $\tilde{U} = \log U$:
\begin{align*}
    \text{Survival-AUPRC}(F_{\text{LN}(\mu,\sigma^2)},L, U) &= \int_0^1\left[F_{\text{LN}(\mu,\sigma^2)}(U/t) -F_{\text{LN}(\mu,\sigma^2)}(Lt)\right]dt \\
    &= \int_0^1 \left[\FN(\tilde{U} - \log t) - \FN(\tilde{L} + \log t)\right]dt \\
    (\text{substituting } s = \log t) &= \int_{-\infty}^{0} \left[\FN(\tilde{U} - s) - \FN(\tilde{L} + s)\right]e^sds \\
    &= \left[\FN(\tilde{U} - s) - \FN(\tilde{L} + s)\right]e^s \big|_{s=-\infty}^{s=0} \\
    &\quad- \int_{-\infty}^{0} \left[-\fN(\tilde{U} - s) - \fN(\tilde{L} + s)\right]e^sds \\
    &= \left(F_{\mathcal{N}(\mu,\sigma^2)}(\tilde{U}) - F_{\mathcal{N}(\mu,\sigma^2)}(\tilde{L})\right) \\ & \qquad + \int_{-\infty}^{0} \left[\fN(\tilde{U} - s) + \fN(\tilde{L} + s)\right]e^sds \\
    &= \left(F_{\mathcal{N}(\mu,\sigma^2)}(\tilde{U}) - F_{\mathcal{N}(\mu,\sigma^2)}(\tilde{L})\right) \\ & \qquad + \int_{-\infty}^{0} \fN(\tilde{U} - s)e^sds + \int_{-\infty}^0\ \fN(\tilde{L} + s)e^sds \\
    &=  \left(F_{\mathcal{N}(\mu,\sigma^2)}(\tilde{U}) - F_{\mathcal{N}(\mu,\sigma^2)}(\tilde{L})\right) \\ & \qquad +\int_{-\infty}^{0} \frac{1}{\sigma}\phi\left(\frac{\tilde{U} - s - \mu}{\sigma}\right)e^sds + \int_{-\infty}^0\frac{1}{\sigma} \phi\left(\frac{\tilde{L} + s - \mu}{\sigma}\right)e^sds \\
    \left(\text{ substituting } u = \frac{\tilde{U} - s - \mu}{\sigma} \right) &= \left(F_{\mathcal{N}(\mu,\sigma^2)}(\tilde{U}) - F_{\mathcal{N}(\mu,\sigma^2)}(\tilde{L})\right) \\ & \qquad + \int_{\infty}^{\frac{\tilde{U} - \mu}{\sigma}} \frac{1}{\sigma}\phi\left(u\right)e^{\tilde{U} - \sigma u - \mu}(-\sigma)du +\int_{-\infty}^0 \frac{1}{\sigma}\phi\left(\frac{\tilde{L} + s - \mu}{\sigma}\right)e^sds\\
    \left(\text{ substituting } v = \frac{\tilde{L} + s - \mu}{\sigma} \right) &= \left(F_{\mathcal{N}(\mu,\sigma^2)}(\tilde{U}) - F_{\mathcal{N}(\mu,\sigma^2)}(\tilde{L})\right) \\ & \qquad + \int_{\infty}^{\frac{\tilde{U} - \mu}{\sigma}} \frac{1}{\sigma}\phi\left(u\right)e^{\tilde{U} - \sigma u - \mu}(-\sigma)du + \int_{-\infty}^{\frac{\tilde{L} - \mu}{\sigma}} \frac{1}{\sigma}\phi\left(v\right)e^{v\sigma - \tilde{L} + \mu}\sigma dv \\
    &= \left(F_{\mathcal{N}(\mu,\sigma^2)}(\tilde{U}) - F_{\mathcal{N}(\mu,\sigma^2)}(\tilde{L})\right) \\ & \qquad - e^{\tilde{U} - \mu}\int_{\infty}^{\frac{\tilde{U} - \mu}{\sigma}} \phi\left(u\right)e^{- \sigma u}du + e^{-\tilde{L} + \mu}\int_{-\infty}^{\frac{\tilde{L} - \mu}{\sigma}} \phi\left(v\right)e^{v\sigma} dv \\
    \left(\text{using } \int e^{cx}\phi(x)dx = e^{\frac{c^2}{2}}\Phi(x - c)\right) &= \left(F_{\mathcal{N}(\mu,\sigma^2)}(\tilde{U}) - F_{\mathcal{N}(\mu,\sigma^2)}(\tilde{L})\right) \\ & \qquad + \frac{U}{e^{\mu}} \left[ e^{\frac{\sigma^2}{2}} \Phi(u + \sigma)  \right]_{u=\infty}^{u=\frac{\tilde{U} - \mu}{\sigma}}+  \frac{e^\mu}{L}\left[e^{\frac{\sigma^2}{2}}\Phi(v - \sigma)  \right]_{v=-\infty}^{v=\frac{\tilde{L} - \mu}{\sigma}} \\
    &= \left(F_{\mathcal{N}(\mu,\sigma^2)}(\tilde{U}) - F_{\mathcal{N}(\mu,\sigma^2)}(\tilde{L})\right) \\ & \qquad + \frac{U}{e^{\mu}} \left[ e^{\frac{\sigma^2}{2}} \Phi\left(\frac{\tilde{U} - \mu}{\sigma} + \sigma\right)  -e^{\frac{\sigma^2}{2}}\right] +  \frac{e^\mu}{L}\left[e^{\frac{\sigma^2}{2}}\Phi\left(\frac{\tilde{L} - \mu}{\sigma} - \sigma\right) \right] \\
    &= \left(F_{\mathcal{N}(\mu,\sigma^2)}(\tilde{U}) - F_{\mathcal{N}(\mu,\sigma^2)}(\tilde{L})\right) \\ & \qquad + e^{\frac{\sigma^2}{2}} \left[\frac{e^\mu}{L} \Phi\left(\frac{\tilde{L} - \mu}{\sigma} - \sigma\right) +  \frac{U}{e^{\mu}}\left(1-\Phi\left(\frac{\tilde{U} - \mu}{\sigma} + \sigma\right)\right)\right] \\
    &= \Phi_{\mu,\sigma^2}(\log U) - \Phi_{\mu,\sigma^2}(\log L) \\ &\qquad + e^{\frac{\sigma^2}{2}} \left[\frac{e^\mu}{L} \Phi\left(\frac{\log L - \mu}{\sigma} - \sigma\right) +  \frac{U}{e^{\mu}}\Phi\left( - \frac{\log U - \mu}{\sigma} - \sigma \right)\right]. 
\end{align*}

\subsection*{\textsc{E. Survival-AUPRC for log-normal (right-censored)}}
\label{sec:appendix-E}

For a general continuous prediction distribution $F$ with an interval outcome $[L, \infty)$, we define Survival-AUPRC as 
\begin{align*}
    \text{Survival-AUPRC}(F, L) &= \int_0^1\left[1 - F(Lt)\right]dt.
\end{align*}
Specifically for the case of log-normal, where $\Phi$ is the CDF of $\mathcal{N}(0,1)$, and $\tilde{L} = \log L$ (following \hyperref[sec:appendix-D]{Appendix-D}),
\begin{align*}
    \text{Survival-AUPRC}(F_{\text{LN}(\mu,\sigma^2)}, L) &= \int_0^1\left[1 -F_{\text{LN}(\mu,\sigma^2)}(Lt)\right]dt = 1 - \Phi_{\mu,\sigma^2}(\tilde L)  + \frac{e^{\mu+\frac{\sigma^2}{2}}}{L} \Phi\left(\frac{\tilde L - \mu}{\sigma} - \sigma\right). 
\end{align*}

\subsection*{\textsc{F. Survival-AUPRC for log-normal (uncensored)}}
\label{sec:appendix-F}

For a general continuous prediction distribution $F$ with a point outcome $y$, we define Survival-AUPRC
\begin{align*}
    \text{Survival-AUPRC}(F, y) &= \int_0^1\left[F(y/t) - F(yt)\right]dt.
\end{align*}
Specifically for the case of log-normal, where $\Phi$ is the CDF of $\mathcal{N}(0,1)$, and $\tilde{y} = \log y$ (following \hyperref[sec:appendix-D]{Appendix-D}),
\begin{align*}
    \text{Survival-AUPRC}(F_{\text{LN}(\mu,\sigma^2)},y) &= \int_0^1\left[F_{\text{LN}(\mu,\sigma^2)}(y/t) -F_{\text{LN}(\mu,\sigma^2)}(yt)\right]dt \\
    &= e^{\frac{\sigma^2}{2}} \left[\frac{e^\mu}{y} \Phi\left(\frac{\tilde y - \mu}{\sigma} - \sigma\right) +  \frac{y}{e^{\mu}}\Phi\left(- \frac{\tilde y - \mu}{\sigma} - \sigma \right)\right].
\end{align*}

\newpage
\subsection*{\textsc{G. Evaluation as binary outcome}}
\label{sec:appendix-G}

\subsection*{\textsc{H. Individual Patients in Interval-Censored Survival-CRPS Model}}
\label{sec:appendix-H}
\begin{figure*}[h]
  \centering
  \includegraphics[width=\linewidth]{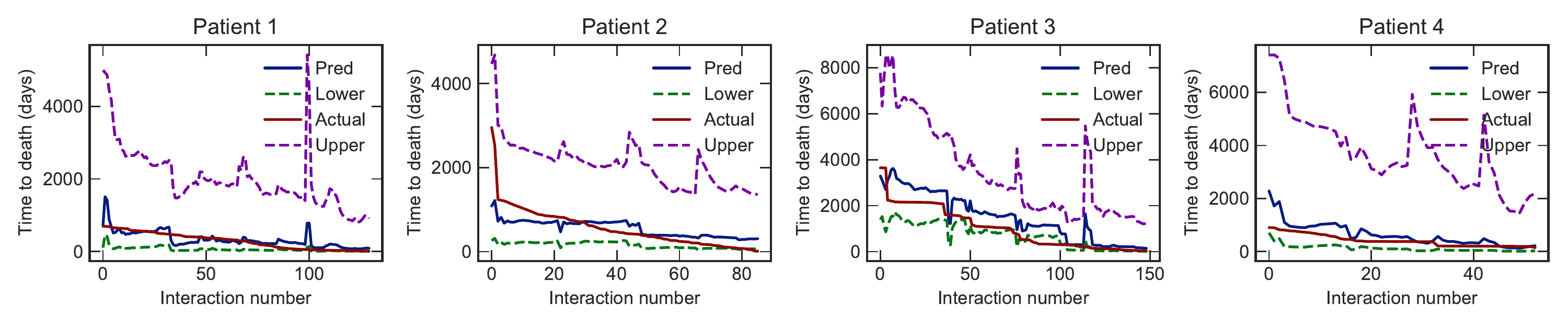}
  \caption{Median predicted time to death (with 95\% intervals) for individual patients from the interval-censored Survival-CRPS model. Our model gives more confident predictions upon repeated interactions between patients and the EHR. True times to death generally lie within predicted intervals.}
  \label{fig-patients}
\end{figure*}

\end{document}